%% file: emnlp-ijcnlp-2019.tex
\documentclass[11pt,a4paper]{article}
\usepackage[hyperref]{emnlp-ijcnlp-2019}
\usepackage{times}
\usepackage{latexsym}
\usepackage{comment}
\usepackage{amsmath}
\usepackage{graphicx}
\usepackage{amssymb}

\DeclareMathOperator*{\argmax}{argmax}

\usepackage{url}

\definecolor{color1}{HTML}{da6752}
\definecolor{color2}{HTML}{5573a6}
\definecolor{color3}{HTML}{6f9f6a}
\definecolor{color4}{HTML}{f3905c}

\aclfinalcopy % Uncomment this line for the final submission

\title{Capturing Greater Context for Question Generation}

\author{{Luu Anh Tuan* ~ Darsh J Shah* ~~~~~~ Regina Barzilay}\\
\mbox{}\\
Computer Science and Artificial Intelligence Lab, MIT \\
\small{\texttt{\{tuanluu, darsh, regina\}@csail.mit.edu}}}

\date{}

\begin{document}

\maketitle
\let\svthefootnote\thefootnote
\let\thefootnote\relax\footnote{Asterisk (\textbf{*}) denotes equal contribution.}
\addtocounter{footnote}{-1}

\begin{abstract}
Automatic question generation can benefit many applications ranging from dialogue systems to reading comprehension.
While questions are often asked with respect to long documents, there are many challenges with modeling such long documents. Many existing techniques generate questions by effectively looking at one sentence at a time, leading to questions that are easy and not reflective of the human process of question generation. Our goal is to incorporate interactions across multiple sentences to generate realistic questions for long documents. In order to link a broad document context to the target answer, we represent the relevant context via a multi-stage attention mechanism, which forms the foundation of a sequence to sequence model. We outperform state-of-the-art methods on question generation on three question-answering datasets -- SQuAD, MS MARCO and NewsQA.

\end{abstract}

\input{sections/00_introduction.tex}
\input{sections/01_related_work.tex}

\input{sections/02_problem_definition.tex}

\input{sections/03_model_architecture.tex}

\input{sections/04_experiments.tex}

\input{sections/05_results.tex}

\input{sections/06_conclusion.tex}

\section{Acknowledgments}
We thank the MIT NLP group for their helpful discussion and comments.
This work is supported by DSO grant DSOCL18002.

\bibliography{emnlp-ijcnlp-2019}
\bibliographystyle{acl_natbib}

\end{document}

%% file: sections/00_introduction.tex
\section{Introduction}

\begin{figure}[h!]
    %\centering
  \includegraphics[width=1\linewidth]{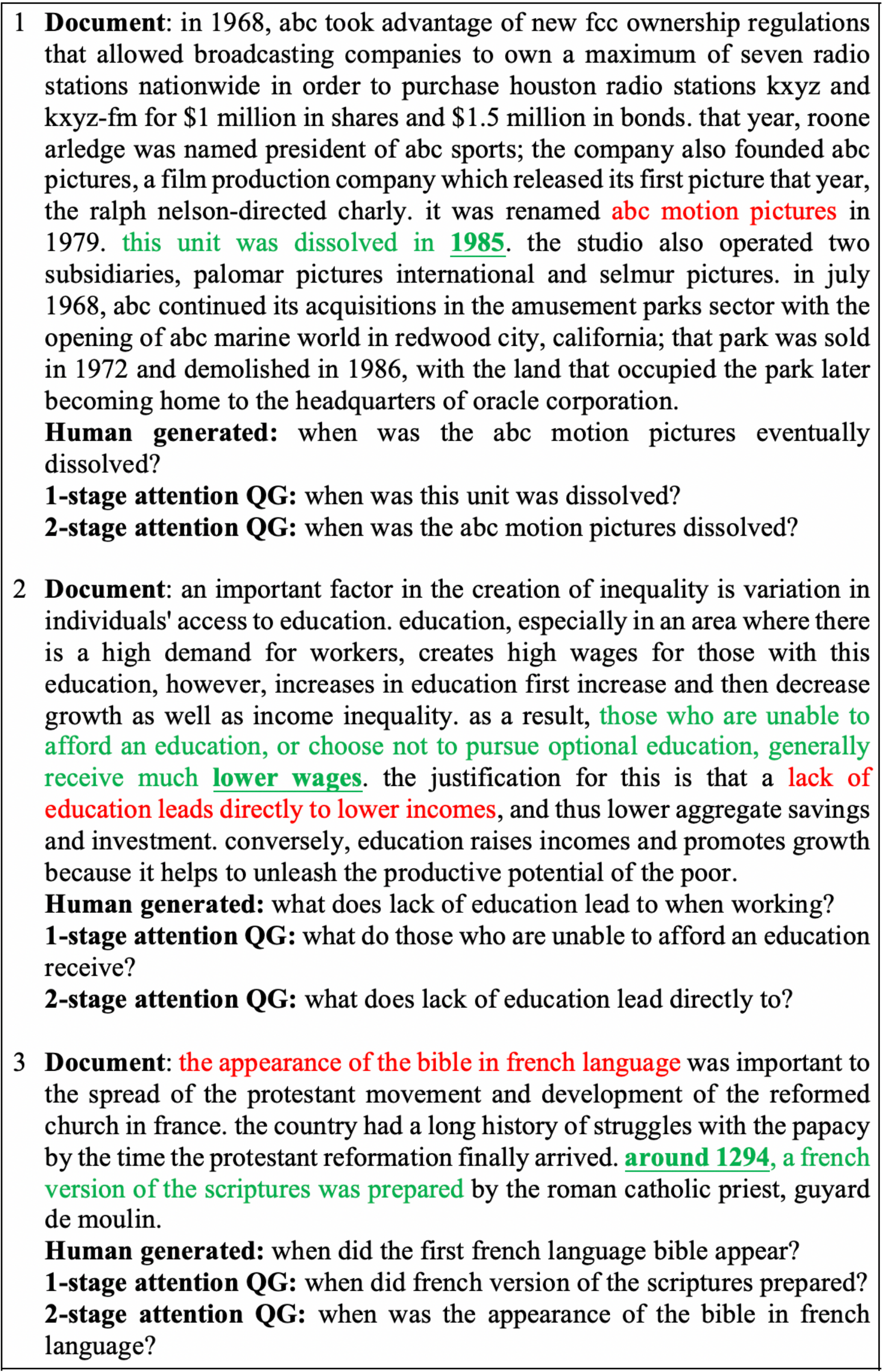}
    \caption{Examples (lower-cased) where multi-sentence context is required to ask the correct questions. Sentences containing answers are in green, while answers are underlined. The red phrases indicate additional background used by a human to generate the question. 1-stage and 2-stage attention QG are results generated by our model with different numbers of attention stages.}
    \label{fig:example}
    \vspace{-3mm}
\end{figure}
 
The tremendous popularity of reading comprehension through datasets like SQuAD \cite{rajpurkar2016}, MS MARCO \cite{nguyen2016} and NewsQA \cite{trischler2016} has led to a surge in machine reading and reasoning techniques. These datasets are typically constructed using crowd sourcing, which provides high quality questions, but at a high cost of manual labor. 
There is an urgent need for automated methods to generate quality question-answer pairs from textual corpora. %These can aid several applications like chatbots, dialogue-systems and assist education, especially in under-developed domains. 

Our goal is to generate a suitable question for a given target answer -- a span of text in a provided document. To this end, we must be able to identify the relevant context for the question-answer pair from the document. Modeling long documents, however, is formidable, and our task involves understanding the relation between the answer and encompassing paragraphs, before asking the relevant question. Typically most existing methods have simplified the task by looking at just the answer containing sentence. 
However, this does not represent the human process of generating questions from a document. For instance, crowd workers for the SQuAD dataset, as illustrated in Figure \ref{fig:example}, used multiple sentences to ask a relevant question. In fact, as pointed out by \cite{Du2017}, around 30\% of the human-generated questions in SQuAD rely on information beyond a single sentence. %Such percentage is even higher in other datasets such as TriviaQA \cite{joshi2017} and HotpotQA \cite{yang2018}which requires more reasoning ability to answer the questions. %Figure \ref{fig:example} shows some examples in SQuAD dataset where the question generation requires context beyond a single sentence.

%Sequence-to-sequence models trained on existing question answering datasets are getting popular, and already outperform rule based methods. 

%Generating questions, as found in these datasets,  requires more context than just the answer sentence. %As pointed out by \cite{Du2017}, around 30\% of the human-generated questions in SQuAD rely on information beyond a single sentence. Such percentage is even higher in other datasets such as TriviaQA \cite{joshi2017} and HotpotQA \cite{yang2018}  which requires more reasoning ability to answer the questions. 
 %Figure \ref{fig:example} shows some examples in SQuAD dataset where the question generation requires context beyond a single sentence. Generating such questions would require an appropriate modelling of the answer and its document.
%Modelling long documents, however, is formidable, and this task involves understanding the relation between the answer and encompassing paragraphs, before asking the right question. %Unlike summarization, the question should not directly capture the answer but utilize its context to implicitly hint and enquire about it. 
%Typically methods have simplified the task by looking at just the answer containing sentence, or feeding large documents as is to a sequence-to-sequence model, or by using syntactic features to gather some answer context.

To accommodate such phenomenon, we propose a novel approach for document-level question generation by explicitly modeling the context based on a multi-stage attention mechanism. %Our framework is able to capture the auxiliary document context \textit{abc motion pictures} from the first example in Figure \ref{fig:example}, while also maintaining the answer's immediate context --- \textit{the unit was dissolved in}. 
As the first step, our method captures the immediate context, by attending the entire document with the answer to highlight phrases, e.g. \textit{``the unit was dissolved in''} from example 1 in Figure \ref{fig:example}, having a direct relationship with the answer, i.e. \textit{``1985''}. In an iterative step thereafter, we attend the original document representation with the attended document computed in the previous step, to expand the context to include more phrases, e.g. \textit{``abc motion pictures''}, that have an indirect relationship with the answer. We can repeat this process multiple times to increase the linkage-level of the answer-related background. 

The final document representation, contains relevant answer context cues by means of attention weights. Through a copy-generate decoding mechanism, where at each step a word is either copied from the input or generated from the vocabulary, the attention weights guide the generation of the context words to produce high quality questions. The entire framework, from context collection to copy-generate style generation is trained end-to-end.

 Our framework for document context representation, strengthened by more attention stages leads to a better question generation quality. Specifically, on SQuAD we get an absolute 5.79 jump in the Rouge points by using a second stage answer-attended representation of the document, compared to directly using the representation right after the first stage. We evaluate our hypothesis of using a controllable context to generate questions on three different QA datasets --- SQuAD, MS MARCO, and NewsQA. Our method strongly outperforms existing state-of-the-art models by an average absolute increase of 1.56 Rouge, 0.97 Meteor and 0.81 Bleu scores over the previous best reported results on all three datasets.

%% file: sections/01_related_work.tex
\section{Related Work}

Question generation has been extensively studied in the past with broadly two main approaches, rule-based and learning-based. 

\textbf{Rule-based techniques} These approaches usually rely on rules and templates of sentences' linguistic structures, and apply heuristics to generate questions \cite{chali2015,Heilman2011,Lindberg2013,labutov2015}. This requires human effort and expert knowledge, making scaling the approach very difficult. Neural methods tend to outperform and generalize better than these techniques. 

%Human written templates and rules for question construction are fairly common \cite{}. These are constrained by the rules, and fail to generalize to other domains, often requiring immense effort in the source domain itself. Encoder-decoder models trained end to end on the corpus of question answering domains \cite{} are becoming increasingly popular \cite{}. 
\textbf{Neural-based models} Since \citet{serban2016,Du2017}, there have been many neural sequence-to-sequence models proposed for question generation tasks. These models
are trained in an end-to-end manner and exploit the corpora of the question answering datasets to outperform rule based methods in many benchmarks. However, in these initial approaches, there is no indication about parts of the document that the decoder should focus on in order to generate the question.

To generate a question for a given answer, \cite{subramanian2017,kim2018,zhou2017,sun2018} applied various techniques to encode answer location information into an annotation vector corresponding to the word positions, thus allowing for better quality answer-focused questions.  \cite{yuan2017}
combined both supervised and reinforcement learning in the training to maximize rewards that measure question quality. \cite{liu2019} presented a syntactic features based method to represent words in the document in order to decide what words to focus on while generating the question.

The above studies, only consider sentence-level question generation, i.e. looking  at  one  document sentence at a time. Recently, \cite{du2018} proposed a method that incorporated coreference knowledge into the neural networks to better encode this linguistically driven connection across entities for document-level question generation. Unfortunately, this work does not capture other relationships like semantic similarity.  As in example 2 of Figure  \ref{fig:example}, two semantic-related phrases ``lower wages" and ``lower incomes" are needed to be linked together to generate the desired question. \cite{zhao2018} proposed another document-level question generation where they apply a gated self-attention mechanism to encode contextual information. However, their self-attention over the entire document is very noisy, redundant and contains many encoded dependencies that are irrelevant.

%% file: sections/02_problem_definition.tex
\section{Problem Definition}

%While the automatic selection of target answers can be classified as a key-phrase extraction problem \cite{du2018} and was extensively studied in NLP community, the problem of generating question, as discussed, still poses a significant challenge.

In this section, we define the task of question generation. Given the document D and the answer A, we are interested in generating the question $\overline{Q}$ that satisfies:

\[\overline{Q} = \argmax_{Q}~Prob(Q|D,A)\]

\noindent where the document $D$ is a sequence of $l_D$ words: 
$D = {\{x_i\}}^{l_D}_{i=1}$
, the answer $A$ of length $l_A$ must be a sub-span of $D$: $A = {\{x_j\}}^{n}_{j=m}$, where $1 \leq m < n \leq l_D $, and the question $\overline{Q}$ is a well-formed sequence of $l_Q$ words: $\overline{Q} = \{y_k\}^{l_Q}_{k=1}$ that can be answered from $D$ using $A$. The generated words $y_k$ can be derived from the document words ${\{x_i\}}^{l_D}_{i=1}$ or from a vocabulary $V$.

%% file: sections/03_model_architecture.tex
\section{Model Architecture}

\begin{figure}[t!]
  \includegraphics[width=1\linewidth]{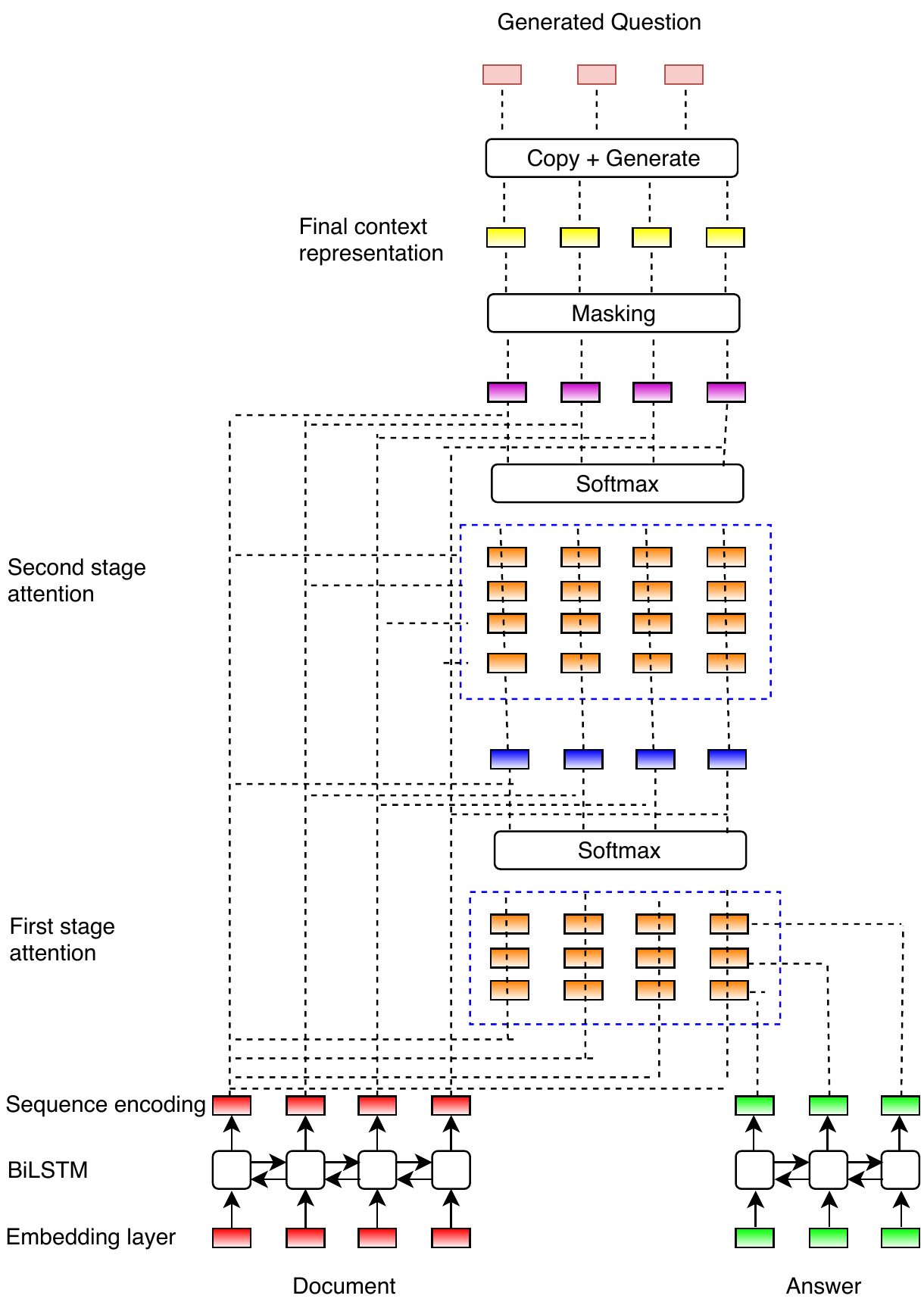}
    \caption{The architecture of our model (with two-stage attention). For simplicity we assume that the document has 4 words and the answer has 3 words.}
    \label{fig:architecture}
\end{figure}

In this section, we describe our proposed model for question generation. The key idea of our model is to use a multi-stage attention mechanism to attend to the important parts of the document that are related to the answer, and use them to generate the question. Figure \ref{fig:architecture} shows the high level architecture of the proposed model.

\subsection{Input and Context Encoding}
The input representation for the document and its interaction with the answer are described as follows.
\label{sec:enc}
\paragraph{Input Encoding}
Our model accepts two inputs, an answer $A$ and the document $D$ that the answer belongs to. Each of which is a sequence of words. The two sequences are indexed into a word embedding layer $W_{emb}$ and then passed into a shared Bidirectional LSTM layer \cite{sak2014long}:
\begin{align}
H^A = \text{BiLSTM}(\mathbf{W_{emb}}(A))\\
H^D = \text{BiLSTM}(\mathbf{W_{emb}}(D))
\end{align}
where $H^A$ $\in$ $\mathbb{R}^{\ell_A \times d}$ and $H^D \in \mathbb{R}^{\ell_D \times d}$ are the hidden representations of $A$ and $D$ respectively, and $d$ is the hidden size of the Bidirectional LSTM.

\paragraph{Context Encoding}

%In order to generate a question that can be answered using the answer, we need to represent the document so that we are capturing the answer and the words related to the answer.
%As a first step, we define the basic context of the answer as the sentence in which it occurs.
%Questions asked from just the sentence in which the answer appears can be quite short and elementary, not capturing the difficulty of questions present in datasets such as SQuad.

%In order to capture the context for question generation, we need to identify important portions of the document related to the answer. We capture this context using multi-stage attention between the answer and the document as follows.\\
The answer's context in the document is identified using our multi-stage attention mechanism, as described below.

\noindent \textbf{Initial Stage} (context with direct relation to answer): 
We pass $H^D,H^A$ into an alignment layer. Firstly, we compute a soft attention affinity matrix between $H^D$ and $H^A$ as follows:
\begin{equation}
    M_{ij}^{(1)} = \textbf{F}(h_{i}^{D})\:\textbf{F}(h_{j}^{A})^{\top} \label{align1}
\end{equation}
where $h_{i}^{D}$ is the $i^{th}$ word in the document and $h_{j}^{A}$ is the $j^{th}$ word in the answer. $\textbf{F}(\cdot)$ is a standard nonlinear transformation function (i.e., $\textbf{F}(x) = \sigma(\textbf{W}x + \textbf{b})$, where $\sigma$ indicates Sigmoid function), and is shared between the document and answer in this stage. $M^{(1)} \in \mathbb{R}^{ \ell_D \times \ell_A }$ is the soft matching matrix. Next, we apply a column-wise max pooling of $M^{(1)}$.
The key idea is to generate an attention vector:
\begin{align}
    a^{(1)} = \text{softmax}(\max_{col}(M^{(1)}))
\end{align}
\noindent where $a^{(1)} \in \mathbb{R}^{~l_D}$. Intuitively, each element $a_i^{(1)} \in a^{(1)}$ captures the degree of relatedness of the $i^{th}$ word in document $D$ to answer $A$ based on its maximum relevance on each word of the answer. To learn the context sensitive weight importance of document, we then apply the attention vector on $H^D$:
\begin{align}
    C^{(1)} = H^D \odot a^{(1)}
\end{align}

\noindent where $\odot$ denotes element-wise multiplication. $C^{(1)} \in R^{l_D \times d}$ can be considered as the first attended contextual representation of document where the words directly related to the answer are amplified with the high attention scores whilst the unrelated words are filtered out with low attention scores.\\

\noindent \textbf{Iterative Stage} (enhance the context with indirect relations): In this stage, we expand the context by collecting more words from the document that are related to \textit{direct-context} computed in the first stage. We achieve this by attending the contextual attention representation of document obtained in stage 1 with original document representation as follows:
\begin{align}
    &M_{ij}^{(2)} = \textbf{F}(h_{i}^{D})\:\textbf{F}(C_{j})^{\top} \\
    &a^{(2)} = \text{softmax}(\max_{\text{col}}(M^{(2)}))\\
    &C^{(2)} = H^D \odot a^{(2)} 
\end{align}

We can repeat the steps in this stage to enhance the context to the answer-related linkage level $k$. We denote the answer-focused context representation after $k$ stages as $C^{(k)}$. In our experiments, we train our models with a predefined value $k$, which is fine-tuned on the validation set.

\paragraph{Answer Masking} Due to the enriched information in the context representation, it is essential for the  model to know the position of the answer so that: (1) it can generate question that is coherent with the answer, and (2) does not include the exact answer in the question. We achieve this by masking the word representation at the position of the answer in the context representation $C^{(k)}$ with a special masking vector:
\begin{align}
    C^{\text{final}} = Mask(C^{(k)})
\end{align}
\noindent $C^{\text{final}} \in R^{l_D \times d}$ can be considered as final contextual attention representation of document and will be used as the input to the decoder.

\subsection{Decoding with Pointer Generator Network}
Using our context rich input representation $C^{\text{final}}$ computed previously, we move forward to the question generation. Our decoding framework is inspired by the pointer-generator network \cite{pointer-generator}. The decoder is a BiLSTM, which at time-step $t$, takes as its input, the word-embedding of the previous time-step's output $W_e(y^{t-1})$ and the latest decoder state attended input representation $r^{t}$ (described later in Equation \eqref{eq:individual_context}) to get the decoder state $h^t$:

\begin{align}
    h^t = BiLSTM([r^t, \mathbf{W_e}(y^{t-1})], h^{t-1})
    \label{eq:decoder}
\end{align}

Using the decoded state to generate the next word, where words can either be copied from the input; or generated by selecting from a fixed vocabulary:
\begin{align}
    P_\text{vocab} = \text{softmax}(\mathbf{V}^{\top}[h^t,r^t])
    \label{eq:fixed_vocabulary}
\end{align}

The \textit{generation probability} $p_\text{gen} \in [0,1]$ at time-step $t$ depends on the context vector $r^t$, the decoder state $h^t$ and the decoder input $x^t = [r^t, \mathbf{W_e}(y^{t-1})]$:
\begin{align}
    p_\text{gen} = \sigma(\mathbf{w_{r}}r^{t} + \mathbf{w_{x}}x^{t} + \mathbf{w_{h}}h^{t})
    \label{eq:generate_copy}
\end{align}

\noindent where $\sigma$ is the sigmoid function. This gating probability $p_\text{gen}$ is used to evaluate the probability of eliciting a word $w$ as follows:
\begin{align}
    P(w) &= p_\text{gen}P_\text{vocab}(w) +  (1-p_\text{gen})\sum_{i:w_{i}=w} a_{i}^{t}
    \label{eq:vocabulary}
\end{align}

\noindent where $\sum_{i:w_{i}=w} a_{i}^{t}$ denotes the probability of word $w$ from the input being generated by the decoder:
\begin{align}
    e_i^{t} &= \mathbf{u}^{\top}\tanh(C^{\text{final}}_i + h^{t-1}) \\
    a^t &= \text{softmax}(e^t) 
\end{align} 
Unlike traditional sequence to sequence models, our input $C^{\text{final}}$ is already weighted via the answer level self-attention. This weighting is reflected directly in the final generation via the copy mechanism through $a^t$, and is also used to evaluate the input context representation $r^t$:
\begin{equation}
    r^t = \sum_{i}a_i^{t}{C^{\text{final}}_i}
    \label{eq:individual_context}
\end{equation}

Finally, the word output at time step $t$, $y^t$ is identified as: 
\begin{align}
    y^{t} &= \argmax_w P(w)
\end{align}

\noindent

The model is trained in an end to end framework, to maximize the probability of generating the target sequence $y^1,...,y^{l_Q}$. At each time step $t$, the probability of predicting $y^t$ is optimized using cross-entropy from the probability of words over the entire vocabulary (fixed and document words). Once the model is trained, we use beam search for inference during decoding. The beam search is parameterised by the possible number of paths $k$.

%% file: sections/04_experiments.tex
\section{Experimental Setup}
In this section we describe the experimental setting to study the proficiency of our proposed model.

\subsection{Datasets}
We evaluate our model on 3 question answering datasets: SQuAD \cite{rajpurkar2016}, MS Marco \cite{nguyen2016} and NewsQA \cite{trischler2016}. These form a comprehensive set of datasets to evaluate question generation.

\vspace{-3mm}

\bigskip
\noindent
\textbf{SQuAD.} SQuAD is a large scale reading comprehension dataset containing close to 100k questions posed by crowd-workers on a set of Wikipedia articles, where the answer is a span in the article. The dataset for our question generation task is constructed from the training and development set of the accessible parts of SQuAD. To be able to directly compare with other reported results, we consider the two following splits: 

\begin{itemize}
    \item Split1: similar to \cite{zhao2018}, we keep the SQuAD train set and randomly split the SQuAD dev set into our dev and test set with the ratio 1:1. The split is done at sentence level.
    \item Split2: similar to \cite{Du2017}, we randomly split the original SQuAD train set randomly into  train and dev set with the ratio 9:1, and keep the SQuAD dev set as our test set.  The split is done at article level.
    
\end{itemize}

\noindent
\textbf{MS MARCO.} MS MARCO is the human developed question answering dataset derived from a million Bing search  queries. Each query is associated with paragraphs from multiple documents resulting from Bing, and the dataset mentions the list of ground truth answers from these paragraphs. Similar to \cite{zhao2018}, we extract a subset of MS Marco where the answers are sub-spans within
the paragraphs, and then randomly split the original train set into train (51k) and dev (6k) sets. We use the 7k questions from the original dev set as our test set.

\vspace{-2mm}
\bigskip
\noindent
\textbf{NewsQA.} NewsQA is the human generated dataset based on CNN news articles. Human crowd-workers are motivated to ask questions from headlines of the articles and the answers are found by other workers from the articles contents.
In our experiment, we select the questions in NewsQA where answers are sub-spans within the articles. As a result, we obtain a dataset with 76k questions for train set, and 4k questions for each dev and test set.

\vspace{-2mm}
\bigskip
\noindent
Table \ref{tab:datasets} gives the details of the three datasets used in our experiments.
\vspace{-2mm}

\begin{table}[h]
\centering
\begin{small}
\begin{tabular}{|c|c|c|c|c|c|c|}
\hline
Dataset & Train & Dev & Test & $l_D$ & $l_Q$ & $l_A$  \\ \hline
SQuAD-1 & 87,488 & 5,267 & 5,272 & 126 & 11 & 3\\
SQuAD-2 & 77,739 & 9,749 & 10,540 & 127 & 11 & 3 \\
MS Marco & 51,000 & 6,000 & 7,000 & 60 & 6 & 15\\
NewsQA & 76,560 & 4,341 & 4,292 & 583  & 8 & 5\\ 
\hline
\end{tabular}
\vspace{-1mm}
\end{small}
\caption{Description of the evaluation datasets. $l_D$ , $l_Q$ and $l_A$ stand for average length of document, question and answer respectively.}
\vspace{-2mm}
\label{tab:datasets}
\end{table}

%The questions in this dataset are more challenging and have ~\TODO{give some dataset metric to quantify this}. ~\TODO{Give the exact number of questions used for train/dev/test}.

\begin{table*}[ht!]
\centering
\begin{tabular}{|l||c|c|c|c||c|c|}
\hline
Model & Bleu-1 & Bleu-2 & Bleu-3 & Bleu-4 & Meteor & Rouge-L  \\ \hline
PCFG-Trans & 28.77 & 17.81 & 12.64 & 9.47 & 18.97 & 31.68 \\
SeqCopyNet & - & - & - & 13.02 & - & 44.00 \\
seq2seq+z+c+GAN & 44.42 & 26.03 & 17.60 & 13.36 & 17.70 & 40.42  \\
NQG++ & 42.36 & 26.33 & 18.46 & 13.51 & 18.18 & 41.60  \\
MPQG  & -       & -      & -  & 13.91 & -   & - \\ 
APM & 43.02 & 28.14 & 20.51 & 15.64 & - & - \\
ASs2s & -      & -         & -   & 16.17 & -  & - \\ 
S2sa-at-mp-gsa & 45.69       & 30.25      & 22.16  & 16.85 & 20.62   & 44.99   \\ 
CGC-QG & 46.58 & 30.90 & 22.82 & 17.55 & 21.24 & 44.53 \\ \hline
Our model & \textbf{46.60}     & \textbf{31.94} & \textbf{23.44} & \textbf{17.76} & \textbf{21.56} & \textbf{45.89} \\ \hline
\end{tabular}
\vspace{-2mm}
\caption{Results in question generation on SQuAD split1}
\label{tab:split1}
\end{table*}

\begin{table*}[ht!]
\centering
\begin{tabular}{|l||c|c|c|c||c|c|}
\hline
Model & Bleu-1 & Bleu-2 & Bleu-3 & Bleu-4 & Meteor & Rouge-L  \\ \hline

LTA & 43.09 & 25.96 & 17.50 & 12.28 & 16.62  & 39.75 \\
MPQG  & -       & -      & -  & 13.98 & 18.77   & 42.72 \\ 
CorefNQG & - & - & 20.90 & 15.16 & 19.12 & - \\
ASs2s & -      & -         & -   & 16.20 & 19.92  & 43.96 \\ 
S2sa-at-mp-gsa & 45.07       & 29.58      & 21.60  & 16.38 & 20.25   & 44.48   \\ \hline
Our model & \textbf{45.13}     & \textbf{30.44}      & \textbf{23.40} & \textbf{17.09} & \textbf{21.25} & \textbf{45.81} \\ \hline
\end{tabular}
\vspace{-2mm}
\caption{Results in question generation on SQuAD split2}
\label{tab:split2}
\vspace{-4mm}
\end{table*}

\subsection{Implementation Details}

 We use a one-layer Bidirectional LSTM with hidden dimension size of 512 for the encoder and decoder. Our entire model is trained end-to-end, with batch size 64, maximum of 200k steps, and Adam optimizer with a learning rate of 0.001 and L2 regularization set to $10^{-6}$. We initialize our word embeddings with frozen pre-trained  GloVe  vectors \cite{Pennington2014}. Text is lowercased and tokenized  with  NLTK. We tune the step of biattention used in encoder from \{1, 2, 3\} on the development set. During decoding, we used beam search with the beam size of 10, and
stopped decoding when every beam in the stack generates the $\textless EOS \textgreater$ token.

\subsection{Evaluation}
Most of the prior studies evaluate the model performances against target questions using automatic metrics. In order to have an empirical comparison, we too use
Bleu-1, Bleu-2, Bleu-3, Bleu-4 \cite{Papineni2002}, METEOR \cite{Denkowski2014} and ROUGE-L \cite{Lin2004} to evaluate the question generation methods.  Bleu
measures the average n-gram precision on a set of reference sentences. METEOR is a recall-oriented metric used to calculate the similarity between generations and references. ROUGE-L is used to evaluate longest common sub-sequence recall of the generated sentences compared to references.  A question structurally and syntactically similar to the human question would score high on these metrics, indicating relevance to the document and answer.

In order to have a more complete evaluation, we also report human evaluation results, where annotators evaluate the quality of questions generated on two important parameters: naturalness (grammar) and difficulty (in answering the question) (Section 6.2).

\begin{table*}[ht!]
\centering
\begin{tabular}{|l||c|c|c|c||c|c|}
\hline
Model & Bleu-1 & Bleu-2 & Bleu-3 & Bleu-4 & Meteor & Rouge-L  \\ \hline
LTA & - & - & - &10.46 & - & - \\
QG+QA & -      & -         & -   & 11.46 & -  & - \\ 
S2sa-at-mp-gsa  & -       & -      & -  & 17.24 & -   & - \\ \hline
Our model & \textbf{41.43} & \textbf{29.97} & \textbf{23.01} & \textbf{18.25} & \textbf{42.77} & \textbf{19.43} \\ \hline
\end{tabular}
\caption{Results in question generation on MS MARCO}
\label{tab:ms_macro}
\end{table*}

\begin{table*}[ht!]
\centering
\begin{tabular}{|l||c|c|c|c||c|c|}
\hline
Model & Bleu-1 & Bleu-2 & Bleu-3 & Bleu-4 & Meteor & Rouge-L  \\ \hline
PCFG-Trans & 16.90 & 7.94 & 4.72 & 3.08 & 13.74 & 23.78 \\
MPQG & 35.70 & 17.16 & 9.64 & 5.65 & 14.13 & 39.85  \\
NQG++ & 40.33 & 22.47 & 14.83 & 9.94 & 16.72 & 42.25 \\
CGC-QG & 40.45 & 23.52 & 15.68 & 11.06 & 17.43 & 43.16 \\ \hline
%Our’s & \textbf{42.13} & \textbf{25.52} & \textbf{17.30} & \textbf{12.27} & \textbf{43.86} & \textbf{19.24} \\ \hline
Our model & \textbf{42.54} & \textbf{26.14} & \textbf{17.30} & \textbf{12.36} & \textbf{19.04} & \textbf{44.05} \\ \hline
\end{tabular}
\caption{Results in question generation on NewsQA}
\label{tab:news_qa}
\end{table*}

\subsection{Baselines}
As baselines, we compare our proposed model against several prior work on question generation. These include:\vspace{-3mm}

\begin{itemize}
\itemsep-0.2em
    \item \textbf{PCFG-Trans} \cite{Heilman2011}: a rule-based system that generates a question based on a given answer word span.
    \item \textbf{LTA} \cite{Du2017}: the seminal Seq2seq model for question generation.
    \item \textbf{ASs2s} \cite{kim2018}: a Seq2Seq model  learns to identify which interrogative word should be used by replacing the answer in the original passage with a special token.
    \item \textbf{MPQG} \cite{Song2018}: a Seq2Seq model that matches the answer with the passage before generating question
    \item \textbf{QG+QA} \cite{duan2017}: a model that combines supervised and reinforcement learning for question generation
    \item \textbf{NQG++} \cite{zhou2017}: a Seq2Seq model with a feature-rich encoder to encode answer position, POS and NER tag information.
    \item \textbf{APM} \cite{sun2018}: a model that incorporates the relative distance between the context words and answer when generating the question.
    \item \textbf{S2sa-at-mp-gsa} \cite{zhao2018} : a Seq2Seq model that uses gate self-attention and maxout-pointer mechanism to encode the context of question.
    \item \textbf{SeqCopyNet} \cite{zhou2018_seq}: a Seq2Seq model that use the copying mechanism to copy not only a single word but a sequence from the input sentence.
    \item \textbf{Seq2seq+z+c+GAN} \cite{yao2018}: a GAN-based model captures the diversity and learning representation using the observed variables.
    \item \textbf{CorefNQG} \cite{du2018}: a Seq2Seq model that utilizes the coreference information to link the contexts.
    \item \textbf{CGC-QG} \cite{liu2019}: a Seq2Seq model that learns to make decisions on which words to generate and to copy using rich syntactic features.
\end{itemize}

%For \textbf{PCFG-Trans}, \textbf{MPQG} and \textbf{NQG++} models,  the results are from experiments conducted using published code on GitHub. For other models, we report results from their papers as is.

\begin{comment}
\bigskip
\noindent
\textbf{Du}: This is the first large scale data driven model. Here the paragraph and question are passed separately to the generate the question. There is no explicit mechanism to focus on this large input while generating a question.

\bigskip
\noindent
\textbf{Yao}:Here they first compute answer encoded document representation by encoding the answer in the paragraph. Then, using self attention, a self-matched representation is presented to the encoder along with the original answer encoded paragraph representation. This is fed to the decoder for final output generation.

\bigskip
\noindent
\textbf{Liu}: Construct a clue word predictor, by computing a dependency tree over the document and running it through a graph convolution. This clue-word predictor, along with other syntactic features is fed in to the model to generate a question by deciding to either generate or copy a word from the passage, guided by these features.

\bigskip
\noindent
\textbf{Masking}: We study the impact of including the masking the answer in the paragraph representation.

\bigskip
\noindent
\textbf{Depth}: We study the impact of varying the depth of the biattention recursion on the generation output.
\end{comment}

%% file: sections/05_results.tex
\section{Results and Analysis}

In this section, we discuss the experimental results and some ablation studies of our proposed model.

\subsection{Comparison with Baseline Models}
We present the question generation performance of baseline models and our model on the three QA datasets in Tables \ref{tab:split1}, \ref{tab:split2}, \ref{tab:ms_macro} and \ref{tab:news_qa} \footnote{For most baselines, we don't have access to their implementations. Hence, we present results for only datasets that they report on in their papers.}. We find that our model consistently outperforms all other baselines and sets a new state-of-the-art on all datasets and across different splits.

For SQuAD split-1, we achieve an average absolute improvement of 0.2 in Bleu-4, 0.3 in  Meteor and 1.3 points in Rouge-L score compared to the best previous reported result. \footnote{We take 5 random splits and report the average across the splits. The lowest performance of the 5 runs also exceeds the state-of-the-art in this setting. Previous methods take an equal random split of the development set into dev/test sets. This can lead to inconsistencies in comparisons.} For SQuAD split-2, we achieve even higher average absolute improvement of 0.7, 1.0 and 1.4 points of Bleu-4, Meteor and Rouge-L scores respectively, compared to \textit{S2sa-at-mp-gsa} - the best previous model on the dataset and also a document-level question generation model. Showing that our model can identify better answer-related context for question generation compared to other document-level methods.  On the MS MARCO dataset, where the ground truth questions are more natural, we achieve an absolute improvement of 1.0 in Bleu-4 score compared to the best previous reported result.

On the NewsQA dataset, which is the harder dataset as the length of input documents are very long, our overall performance is still promising. Our model outperforms the CGC-QG model by an average absolute score 1.3  of Bleu-4, 1.6 of Meteor, and 0.9 of Rouge-L, again demonstrating that exploiting the broader context can help the question generation system better match humans at the task.

\subsection{Human Evaluation}

To measure the quality of questions generated by our system, we conduct a human evaluation. Most of the previous work, except the LTA system \cite{Du2017}, do not conduct any human evaluation, and for most of the competing methods, we do not have the code to reproduce the outputs. Hence, we conduct human evaluation using the exact same settings and metrics in \cite{Du2017} for a fair comparison. Specifically, we consider two criterion in human evaluation: (1) Naturalness, which indicates the grammaticality and fluency; and (2) Difficulty, which measures the syntactic divergence and the reasoning needed to answer the question. We randomly sample 100 sentence-question pairs from our SQuAD experimental outputs. We then ask four professional English speakers to rate the pairs in terms of the above criterion on a 1$-$5 scale (5 for the best). The experimental result is given in Table \ref{tab:human_evaluation}.

\begin{table}[h]
\centering
\begin{small}
\begin{tabular}{|p{3.5cm}|p{1.5cm}|p{1.5cm}|}
\hline
 &  Naturalness & Difficulty  \\ \hline
LTA  & 3.36 & 3.03\\
Our model & 3.68 & \textbf{3.27} \\ \hline
Human generated questions & \textbf{4.06} & 2.65\\
\hline
\end{tabular}
\end{small}
\caption{Human evaluation results for question
generation. Naturalness and difficulty are rated
on a 1$-$5 scale (5 for the best).}
\label{tab:human_evaluation}
\end{table}
The inter-rater agreement of Krippendorff's Alpha between human evaluations is 0.21. The results imply that our model can generate questions of better quality than the LTA system. Our system tends to generate difficult questions owing to the fact that it gathers context from the whole document rather than from just one or two sentences.

\subsection{Ablation Study}

In this section, we study the impact of (1) The proposed attention mechanism in the encoder; (2) The number of attention stages used in that mechanism; and (3) The masking technique used for the encoder.

\begin{table*}[h!]
\centering
%\begin{small}
\begin{tabular}{|p{5.0cm}||p{1.1cm}|p{1.1cm}|p{1.5cm}|}
\hline
Model &  Bleu-4 & Meteor & Rouge-L  \\ \hline
\noindent Original (2-stage attention) & \textbf{17.76} & \textbf{21.56} & \textbf{45.89} \\ 
~~~ - without attention & 3.06 & 10.83 &  28.75\\
~~~ - without masking & 5.19 & 13.08 & 31.14\\
~~~ - with 1-stage attention & 14.52 & 18.28 & 40.10 \\
~~~ - with 3-stage attention & 12.87 & 16.05 & 38.33 \\
\hline
\end{tabular}
%\end{small}
\caption{Ablation study on SQuAD split 1.}
\label{tab:ablation}
\end{table*}

\bigskip
\noindent
\textbf{Impact of using encoder attention~} In this ablation, we remove the attention mechanism in the encoder and just pass the vanilla document representation to the decoder. As shown in Table \ref{tab:ablation}, without using attention mechanism, the performance drops significantly (more than 14 Bleu points). We hypothesize that without attention, the model lacks the capability to identify the import parts of document and hence generates questions unrelated to the target answer.

\bigskip
\noindent
\textbf{Impact of number of attention stages~} As shown in Table \ref{tab:ablation}, with an increase in the number of attention stages from 1 to 2, the performance of model improves significantly, with an increment of more than 3 Bleu-4 points. 

\begin{figure}[t!]
    \centering
  \includegraphics[width=0.8\linewidth]{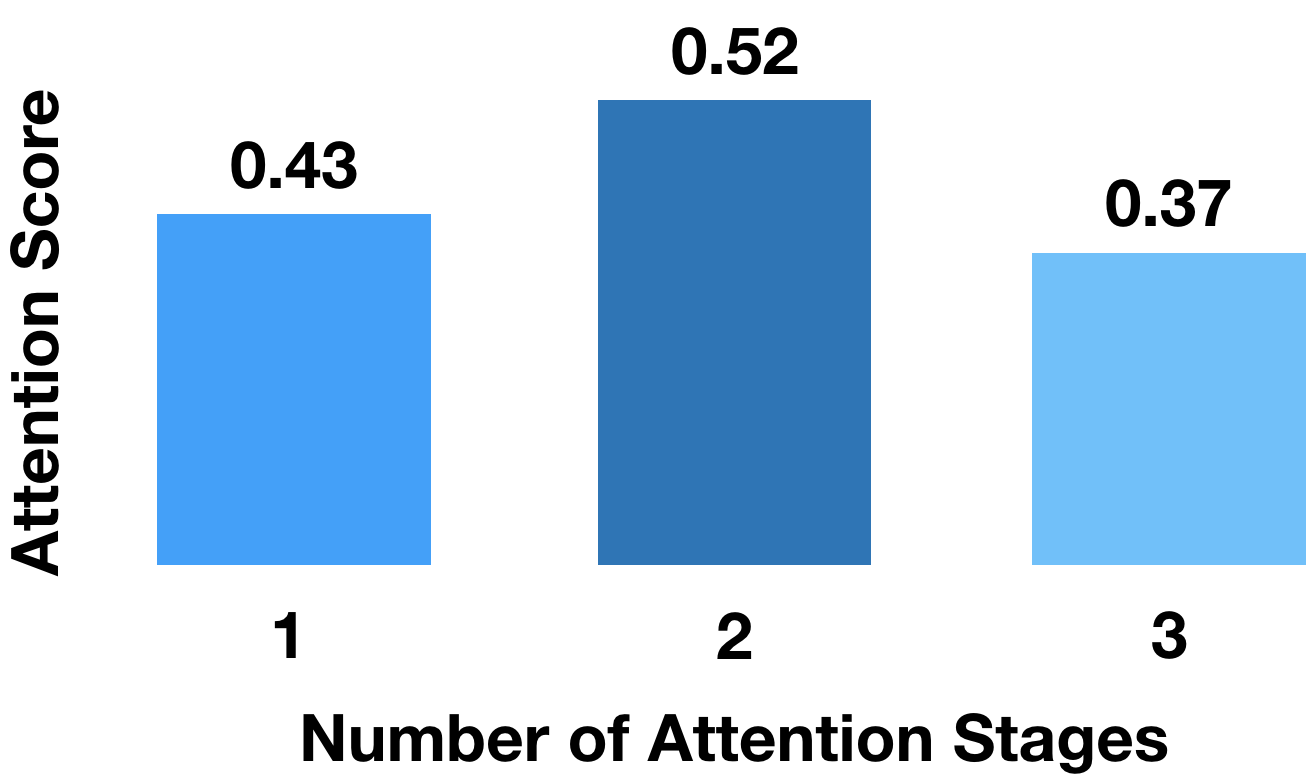}
    \caption{Average total attention score of words in the document that occurred in the ground truth question when using different attention stages (SQuAD split 1).}
    \label{fig:score}
    \vspace{2mm}
\end{figure}

\begin{figure}[t!]
    %\centering
  \includegraphics[width=1.0\linewidth]{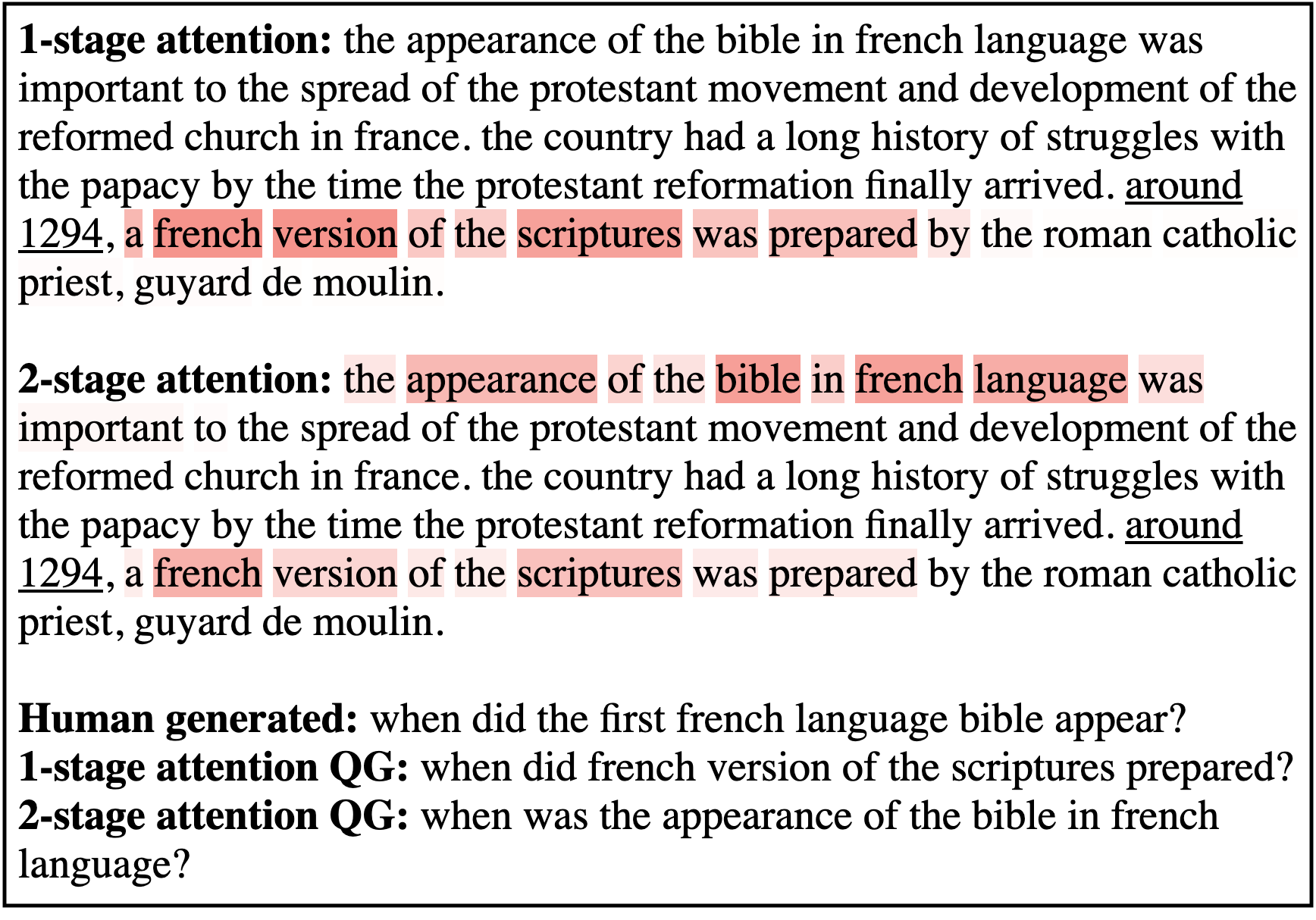}
    \caption{Qualitative analysis of attention vector. The intensity of
the color (red) denotes the strength of the attention weights.}
    \label{fig:density}
\end{figure}

To have a deeper understanding about the impact of the number of attention stages, we calculate for the words in the document that occurred in the ground truth question, their total attention score at the end of input attention layer as in Figure \ref{fig:score}. For 1-stage and 2-stage attention, the total attention score of the question words to be copied from the document are 0.43 and 0.52 respectively, demonstrating that in SQuAD dataset, the 2-stage attention covers more of the question words in a focused manner. An example for this effect can be seen in Figure \ref{fig:density}. The extra stage clearly helps in gathering more relevant context to generate a question closer to the ground-truth.

However, on further increasing the number of attention stages to 3, we observe that the quality of generated questions deteriorates. This can be attributed to the fact that for most of the questions in SQuAD, such 3-stage attention leads to a very cloudy context, where several words get covered, but with a diluted attention. 3-stage attention's coverage in Figure \ref{fig:score} shows this clearly, where its coverage in ground-truth questions is lower than even the 1-stage attention, justifying its poor question generation quality.

\bigskip
\noindent
\textbf{Impact of masking~} While attending to the answer's context and the related sentences is crucial, we find that it is imperative to mask out the answer before getting the input representation. It is demonstrated from the experimental results in Table \ref{tab:ablation} where the Bleu-4 score is increased more than 12 points when applying this masking. %The masking helps in removing the trivial situation of the answer having a high attention score with itself, thus increasing its probability of getting copied. Copying pieces of the answer has poor consequences, and negates the aim of having a question being difficult to be answered. %It can be explained as the answer having a high attention score with itself, leads it to have a higher probability for being copied by the decoder. 

\subsection{Case Study}

In Figure \ref{fig:example}, we introduce some examples that the document-level information obtained from our proposed multi-stage attention mechanism is needed to generate the correct questions.

In example 1, the two-stage attention model is able to identify the phrases \textit{``this unit''} referring to \textit{``abc motion pictures''}, which is out of the sentence containing the answer.

In example 2, two semantic-related phrases \textit{``lower incomes''} and \textit{``lower wages''}  in two different sentences are successfully linked by our two-stage attention model to generate the correct question.

In example 3, the two-stage attention model is able to link two different sentences containing the same word (\textit{``french''}) and semantic-related words (\textit{``bible''} and \textit{``scriptures''}), forming relevant context for generating the expected question.

%% file: sections/06_conclusion.tex
\section{Conclusion}

In this paper, we proposed a novel document-level approach for question generation by using multi-step recursive attention mechanism on the document and answer representation to extend the relevant context. We demonstrate that taking additional attention steps helps learn a more relevant context, leading to a better quality of generated questions. We evaluate our method on three QA datasets - SQuAD, MS MARCO and NewsQA, and set the new state-of-the-art results in question generation for all of them.

%we develop a novel method to recursively gather answer-related context, by repeatedly attending it to the document and utilize this context attended document representation to generate questions. We demonstrate that taking repeated attention steps  in this direction helps learn a focused and more relevant correspondence, leading to a better quality of generated questions. We evaluate our method over three QA datasets - SQuAD, MS Marco and  NewsQA, and achieve state-of-the-art results in question generation on all of those datasets.

%% file: emnlp-ijcnlp-2019.bbl
\begin{thebibliography}{27}
\expandafter\ifx\csname natexlab\endcsname\relax\def\natexlab#1{#1}\fi

\bibitem[{Chali and Hasan(2015)}]{chali2015}
Yllias Chali and Sadid~A Hasan. 2015.
\newblock Towards topic-to-question generation.
\newblock \emph{Computational Linguistics}, 41(1):1--20.

\bibitem[{Denkowski and Lavie(2014)}]{Denkowski2014}
Michael Denkowski and Alon Lavie. 2014.
\newblock Meteor universal: Language specific translation evaluation for any
  target language.
\newblock In \emph{Proceedings of the ninth workshop on statistical machine
  translation}, pages 376--380.

\bibitem[{Du and Cardie(2018)}]{du2018}
Xinya Du and Claire Cardie. 2018.
\newblock Harvesting paragraph-level question-answer pairs from wikipedia.
\newblock \emph{arXiv preprint arXiv:1805.05942}.

\bibitem[{Du et~al.(2017)Du, Shao, and Cardie}]{Du2017}
Xinya Du, Junru Shao, and Claire Cardie. 2017.
\newblock Learning to ask: Neural question generation for reading
  comprehension.
\newblock \emph{arXiv preprint arXiv:1705.00106}.

\bibitem[{Duan et~al.(2017)Duan, Tang, Chen, and Zhou}]{duan2017}
Nan Duan, Duyu Tang, Peng Chen, and Ming Zhou. 2017.
\newblock Question generation for question answering.
\newblock In \emph{Proceedings of the 2017 Conference on Empirical Methods in
  Natural Language Processing}, pages 866--874.

\bibitem[{Heilman(2011)}]{Heilman2011}
Michael Heilman. 2011.
\newblock Automatic factual question generation from text.

\bibitem[{Kim et~al.(2018)Kim, Lee, Shin, and Jung}]{kim2018}
Yanghoon Kim, Hwanhee Lee, Joongbo Shin, and Kyomin Jung. 2018.
\newblock Improving neural question generation using answer separation.
\newblock \emph{arXiv preprint arXiv:1809.02393}.

\bibitem[{Labutov et~al.(2015)Labutov, Basu, and Vanderwende}]{labutov2015}
Igor Labutov, Sumit Basu, and Lucy Vanderwende. 2015.
\newblock Deep questions without deep understanding.
\newblock In \emph{Proceedings of the 53rd Annual Meeting of the Association
  for Computational Linguistics and the 7th International Joint Conference on
  Natural Language Processing (Volume 1: Long Papers)}, volume~1, pages
  889--898.

\bibitem[{Lin(2004)}]{Lin2004}
Chin-Yew Lin. 2004.
\newblock Rouge: A package for automatic evaluation of summaries.
\newblock \emph{Text Summarization Branches Out}.

\bibitem[{Lindberg et~al.(2013)Lindberg, Popowich, Nesbit, and
  Winne}]{Lindberg2013}
David Lindberg, Fred Popowich, John Nesbit, and Phil Winne. 2013.
\newblock Generating natural language questions to support learning on-line.
\newblock In \emph{Proceedings of the 14th European Workshop on Natural
  Language Generation}, pages 105--114.

\bibitem[{Liu et~al.(2019)Liu, Zhao, Niu, Lai, He, Wei, and Xu}]{liu2019}
Bang Liu, Mingjun Zhao, Di~Niu, Kunfeng Lai, Yancheng He, Haojie Wei, and
  Yu~Xu. 2019.
\newblock Learning to generate questions by learning what not to generate.
\newblock \emph{arXiv preprint arXiv:1902.10418}.

\bibitem[{Nguyen et~al.(2016)Nguyen, Rosenberg, Song, Gao, Tiwary, Majumder,
  and Deng}]{nguyen2016}
Tri Nguyen, Mir Rosenberg, Xia Song, Jianfeng Gao, Saurabh Tiwary, Rangan
  Majumder, and Li~Deng. 2016.
\newblock Ms marco: A human generated machine reading comprehension dataset.
\newblock \emph{arXiv preprint arXiv:1611.09268}.

\bibitem[{Papineni et~al.(2002)Papineni, Roukos, Ward, and Zhu}]{Papineni2002}
Kishore Papineni, Salim Roukos, Todd Ward, and Wei-Jing Zhu. 2002.
\newblock Bleu: a method for automatic evaluation of machine translation.
\newblock In \emph{Proceedings of the 40th annual meeting on association for
  computational linguistics}, pages 311--318. Association for Computational
  Linguistics.

\bibitem[{Pennington et~al.(2014)Pennington, Socher, and
  Manning}]{Pennington2014}
Jeffrey Pennington, Richard Socher, and Christopher Manning. 2014.
\newblock Glove: Global vectors for word representation.
\newblock In \emph{Proceedings of the 2014 conference on empirical methods in
  natural language processing (EMNLP)}, pages 1532--1543.

\bibitem[{Rajpurkar et~al.(2016)Rajpurkar, Zhang, Lopyrev, and
  Liang}]{rajpurkar2016}
Pranav Rajpurkar, Jian Zhang, Konstantin Lopyrev, and Percy Liang. 2016.
\newblock Squad: 100,000+ questions for machine comprehension of text.
\newblock \emph{arXiv preprint arXiv:1606.05250}.

\bibitem[{Sak et~al.(2014)Sak, Senior, and Beaufays}]{sak2014long}
Ha{\c{s}}im Sak, Andrew Senior, and Fran{\c{c}}oise Beaufays. 2014.
\newblock Long short-term memory recurrent neural network architectures for
  large scale acoustic modeling.
\newblock In \emph{Fifteenth annual conference of the international speech
  communication association}.

\bibitem[{See et~al.(2017)See, Liu, and Manning}]{pointer-generator}
Abigail See, Peter~J. Liu, and Christopher~D. Manning. 2017.
\newblock \href {https://doi.org/10.18653/v1/P17-1099} {Get to the point:
  Summarization with pointer-generator networks}.
\newblock In \emph{Proceedings of the 55th Annual Meeting of the Association
  for Computational Linguistics (Volume 1: Long Papers)}, pages 1073--1083,
  Vancouver, Canada. Association for Computational Linguistics.

\bibitem[{Serban et~al.(2016)Serban, Garc{\'\i}a-Dur{\'a}n, Gulcehre, Ahn,
  Chandar, Courville, and Bengio}]{serban2016}
Iulian~Vlad Serban, Alberto Garc{\'\i}a-Dur{\'a}n, Caglar Gulcehre, Sungjin
  Ahn, Sarath Chandar, Aaron Courville, and Yoshua Bengio. 2016.
\newblock Generating factoid questions with recurrent neural networks: The 30m
  factoid question-answer corpus.
\newblock \emph{arXiv preprint arXiv:1603.06807}.

\bibitem[{Song et~al.(2018)Song, Wang, Hamza, Zhang, and Gildea}]{Song2018}
Linfeng Song, Zhiguo Wang, Wael Hamza, Yue Zhang, and Daniel Gildea. 2018.
\newblock Leveraging context information for natural question generation.
\newblock In \emph{Proceedings of the 2018 Conference of the North American
  Chapter of the Association for Computational Linguistics: Human Language
  Technologies, Volume 2 (Short Papers)}, pages 569--574.

\bibitem[{Subramanian et~al.(2017)Subramanian, Wang, Yuan, Zhang, Bengio, and
  Trischler}]{subramanian2017}
Sandeep Subramanian, Tong Wang, Xingdi Yuan, Saizheng Zhang, Yoshua Bengio, and
  Adam Trischler. 2017.
\newblock Neural models for key phrase detection and question generation.
\newblock \emph{arXiv preprint arXiv:1706.04560}.

\bibitem[{Sun et~al.(2018)Sun, Liu, Lyu, He, Ma, and Wang}]{sun2018}
Xingwu Sun, Jing Liu, Yajuan Lyu, Wei He, Yanjun Ma, and Shi Wang. 2018.
\newblock Answer-focused and position-aware neural question generation.
\newblock In \emph{Proceedings of the 2018 Conference on Empirical Methods in
  Natural Language Processing}, pages 3930--3939.

\bibitem[{Trischler et~al.(2016)Trischler, Wang, Yuan, Harris, Sordoni,
  Bachman, and Suleman}]{trischler2016}
Adam Trischler, Tong Wang, Xingdi Yuan, Justin Harris, Alessandro Sordoni,
  Philip Bachman, and Kaheer Suleman. 2016.
\newblock Newsqa: A machine comprehension dataset.
\newblock \emph{arXiv preprint arXiv:1611.09830}.

\bibitem[{Yao et~al.(2018)Yao, Zhang, Luo, Tao, and Wu}]{yao2018}
Kaichun Yao, Libo Zhang, Tiejian Luo, Lili Tao, and Yanjun Wu. 2018.
\newblock Teaching machines to ask questions.
\newblock In \emph{IJCAI Conference}.

\bibitem[{Yuan et~al.(2017)Yuan, Wang, Gulcehre, Sordoni, Bachman, Subramanian,
  Zhang, and Trischler}]{yuan2017}
Xingdi Yuan, Tong Wang, Caglar Gulcehre, Alessandro Sordoni, Philip Bachman,
  Sandeep Subramanian, Saizheng Zhang, and Adam Trischler. 2017.
\newblock Machine comprehension by text-to-text neural question generation.
\newblock \emph{arXiv preprint arXiv:1705.02012}.

\bibitem[{Zhao et~al.(2018)Zhao, Ni, Ding, and Ke}]{zhao2018}
Yao Zhao, Xiaochuan Ni, Yuanyuan Ding, and Qifa Ke. 2018.
\newblock Paragraph-level neural question generation with maxout pointer and
  gated self-attention networks.
\newblock In \emph{Proceedings of the 2018 Conference on Empirical Methods in
  Natural Language Processing}, pages 3901--3910.

\bibitem[{Zhou et~al.(2017)Zhou, Yang, Wei, Tan, Bao, and Zhou}]{zhou2017}
Qingyu Zhou, Nan Yang, Furu Wei, Chuanqi Tan, Hangbo Bao, and Ming Zhou. 2017.
\newblock Neural question generation from text: A preliminary study.
\newblock In \emph{National CCF Conference on Natural Language Processing and
  Chinese Computing}, pages 662--671. Springer.

\bibitem[{Zhou et~al.(2018)Zhou, Yang, Wei, and Zhou}]{zhou2018_seq}
Qingyu Zhou, Nan Yang, Furu Wei, and Ming Zhou. 2018.
\newblock Sequential copying networks.
\newblock In \emph{Thirty-Second AAAI Conference on Artificial Intelligence}.

\end{thebibliography}
